\documentclass[sigconf,nonacm]{aamas}

\usepackage{balance} 

\setcopyright{ifaamas}
\acmConference[AAMAS '25]{Proc.\@ of the 24th International Conference
on Autonomous Agents and Multiagent Systems (AAMAS 2025)}{May 19 -- 23, 2025}
{Detroit, Michigan, USA}{Y.~Vorobeychik, S.~Das, A.~Nowe (eds.)}
\copyrightyear{2025}
\acmYear{2025}
\acmDOI{}
\acmPrice{}
\acmISBN{}

\acmSubmissionID{565}

\title[AAMAS-2025 Kuzmenko]{Knowledge Transfer in Model-Based Reinforcement Learning Agents for Efficient Multi-Task Learning}

\subtitle{Extended Abstract}

\author{Dmytro Kuzmenko}
\affiliation{
  \institution{National University of Kyiv-Mohyla Academy}
  \city{Kyiv}
  \country{Ukraine}
}
\email{kuzmenko@ukma.edu.ua}

\author{Nadiya Shvai}
\affiliation{
  \institution{National University of Kyiv-Mohyla Academy}
  \city{Kyiv}
  \country{Ukraine}
}
\email{n.shvay@ukma.edu.ua}
\affiliation{
  \institution{Cyclope.ai}
  \city{Paris}
  \country{France}}
\email{nadiya.shvai@cyclope.ai}

\begin{abstract}
We propose an efficient knowledge transfer approach for model-based reinforcement learning, addressing the challenge of deploying large world models in resource-constrained environments. Our method distills a high-capacity multi-task agent (317M parameters) into a compact 1M parameter model, achieving state-of-the-art performance on the MT30 benchmark with a normalized score of 28.45, a substantial improvement over the original 1M parameter model's score of 18.93. This demonstrates the ability of our distillation technique to consolidate complex multi-task knowledge effectively. Additionally, we apply FP16 post-training quantization, reducing the model size by 50\% while maintaining performance. Our work bridges the gap between the power of large models and practical deployment constraints, offering a scalable solution for efficient and accessible multi-task reinforcement learning in robotics and other resource-limited domains.
\end{abstract}


\newcommand{\BibTeX}{\rm B\kern-.05em{\sc i\kern-.025em b}\kern-.08em\TeX}

\begin{document}


\pagestyle{fancy}
\fancyhead{}


\maketitle 


\section{Introduction}

Reinforcement learning (RL) has achieved significant progress in diverse domains, yet it remains challenging to efficiently train agents for multiple tasks in resource-constrained environments. Model-based RL approaches, such as TD-MPC2 \cite{hansen2024td}, leverage large world models for superior performance and generalization but require substantial computational resources, limiting real-world applicability. Our work addresses this by focusing on optimizing large model-based RL agents for efficient multi-task learning through knowledge distillation and model compression. Specifically, we transfer knowledge from high-capacity TD-MPC2 models to compact 1M parameter backbones suitable for deployment in resource-limited scenarios. Our method builds on traditional teacher-student distillation \citeN{hinton2015distilling, rusu2015policy, teh2017distral, parisotto2015actor, czarnecki2019distilling} and incorporates quantization techniques \cite{micikevicius2018mixed}, creating a pipeline for lightweight RL models. Using the MT30 benchmark \citeN{hansen2024td, tassa2020dmcontrol}, our FP16-quantized model achieves state-of-the-art performance (28.45 normalized score), surpassing the original model (+48.5\%) and outperforming models trained from scratch (+2.77\%). By combining extended distillation periods, optimized batch sizes, and quantization, our approach bridges the gap between the impressive capabilities of large-scale RL models and their deployment in robotics and other real-world applications.


\section{Methods}

Our approach employs a teacher-student distillation framework, adapting it to the specific challenges of model-based RL (Figure 1).

The 317M-parameter TD-MPC2 model servers as the teacher, representing the largest available checkpoint from the TD-MPC2 \cite{hansen2024td}, and the respective 1M-parameter checkpoint serves as the student.

We build upon the original TD-MPC2 loss functions (consistency, reward, and value losses) by introducing an additional reward distillation loss. This new loss is computed as the mean squared error (MSE) between the rewards predicted by the teacher and student models:
\[
L_\text{distill} = \text{MSE}(R_\text{teacher}(s, a), R_\text{student}(s, a))
\]
where $R_\text{teacher}$ and $R_\text{student}$ are the reward predictions of the teacher and student models, respectively.

An additional reward distillation coefficient (\textit{d\_coef}) is introduced to balance the original TD-MPC2 loss with our new distillation loss:
\[
total\_loss = original\_loss + d_\text{coef} * distillation\_loss
\]
The \textit{d\_coef} acts as a hyperparameter controlling the influence of the teacher model's knowledge on the student model's learning process. We empirically find that values close to 0.5 yield the best results, with 0.4 being optimal for most training setups (Table 1).

The normalized score is used as our primary evaluation metric, consistent with \cite{hansen2024td}. Each task is scored on a scale of 1 to 1000, with the average sum divided by the number of tasks, resulting in a final range of 1-100.

For training, we take a 317M parameter TD-MPC2 model checkpoint pretrained on the MT30 dataset. We then train the 1M parameter student model using our distillation approach. The student is trained on the same dataset but with the additional reward distillation loss. We explore different distillation periods, from 200,000 steps to 1,000,000 steps, and various batch sizes to understand the impact of extended distillation on performance.

Lastly, FP16 quantization is applied to reduce model size with minimal impact on task performance, highlighting the method's scalability for resource-constrained deployments. 

For the MT30 benchmark, we utilize the full available dataset of 690,000 episodes (345,690,000 transitions). We conduct our experiments on a desktop PC with a single RTX 3060 GPU (12GB VRAM). 

\begin{figure}[ht]
  \centering
  \includegraphics[width=1.0\linewidth]{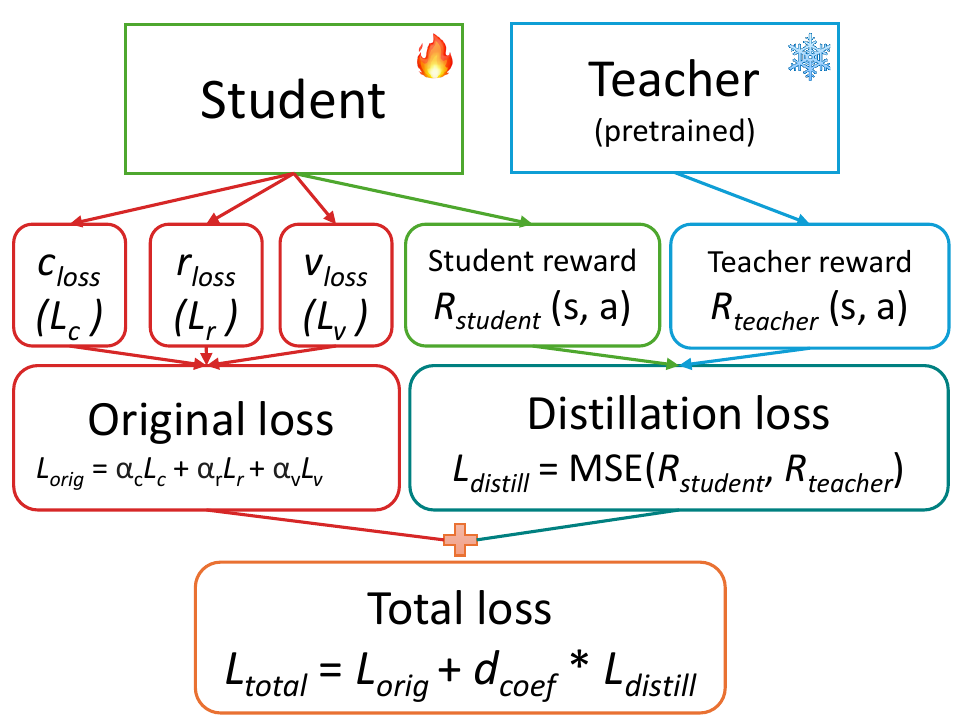}
  \caption{Our distillation approach consists of two main loss function components: the original loss from TD-MPC2 (in red), calculated as a linear combination of consistency, reward, and value losses ($\alpha$ denotes a scaling coefficient for each respective loss); and the distillation loss (in teal) that calculates MSE between student's (green) and teacher's (blue) rewards produced from inferring the same state-action pair. The total loss is a linear combination of both losses, with the distillation loss scaled by the \textit{d\_coef}. In our scenario, the student is trainable while the teacher's weights are frozen.}
  \label{fig:logo}
  \Description{The figure depicts a diagram of our proposed distillation pipeline. A pretrained teacher and a trainable student are used to infer rewards for reward distillation loss calculation. The original loss from TD-MPC2 is preserved, and the total loss is a linear combination of both losses, with the distillation loss scaled by the \textit{d\_coef}.}
\end{figure}

\section{Results}

One impactful finding of our study is the substantial improvement achieved through distillation, particularly with extended training and reduced batch size.

Short-term distillation results showed a significant improvement over the baseline. The distilled model with \textit{d\_coef} = 0.4 achieved a normalized score of 17.85, compared to the baseline score of 14.04, representing a 27.1\% improvement. Extended distillation yielded similar results. The distilled model (\textit{d\_coef} = 0.45) achieved a normalized score of 28.12, compared to the model trained from scratch with a score of 27.36, representing a 2.77\% improvement.

Our findings indicate that for our distillation approach in the context of TD-MPC2, a batch size of 256 offers an optimal balance of performance, efficiency, and hardware compatibility (Table 2).

Comparing distillation from different teacher model sizes revealed the value of using a larger, more knowledgeable teacher model. Distillation from the 48M parameter teacher resulted in a score of 13.61, while distillation from the 317M parameter teacher achieved 17.85, representing a 31.2\% relative improvement.

Our experiments reveal that incorporating next-state latent distillation alongside reward distillation presents a significant challenge due to the dimensional mismatch between teacher (1376-dim) and student (128-dim) models. This mismatch necessitated dimensionality reduction techniques, such as linear projection and PCA, which introduced substantial information loss and led to poor generalization. Linear projection yielded a normalized score of 7.69, while PCA marginally improved this to 8.78, but both fell far short of the 14.04 achieved with reward-only distillation. The additional computational overhead of PCA, combined with its limited gains, further underscored its inefficiency.

These findings suggest that next-state prediction, which relies on abstract intermediate representations, often deviates from the distilled model's optimal structure, making it suboptimal for our setup. In contrast, reward prediction directly aligns with task-specific performance, simplifying the learning process and enabling better outcomes, particularly in reward-oriented tasks like \textit{pendulum-swingup} and \textit{cup-catch}.

\begin{table}[t]
	\caption{Impact of \textit{d\_coef} on student's performance, 200K distillation steps with 317M parameter teacher and batch size of 256.}
	\label{tab:distill coef}
	\begin{tabular}{cc}
            \toprule
		\textit{Distillation coefficient} & \textit{Normalized score} \\ \midrule
		0.05 & 13.61 \\ 
            0.25 & 14.49 \\ 
            \textbf{0.4} & \textbf{17.85} \\
            0.55 & 16.08 \\ 
            0.6 & 14.83 \\ 
            0.9 & 13.79 \\ \bottomrule
	\end{tabular}
\end{table}

\begin{table}[t]
	\caption{Results of different setups of 1M-parameter model distillation vs training from scratch on MT30 benchmark.}
	\label{tab:results}
	\begin{tabular}{lccc}\toprule
		\textit{Method} & \textit{Batch size} & \textit{Training steps, \#} & \textit{Score}\\  \midrule
            distill & 128 & 200K & 17.37 \\
            from scratch & 256 & 200K & 14.04 \\ 
  		distill & 256 & 200K & \textbf{17.85} \\ 
            \hline
            from scratch & 1024 & 200K & \textbf{18.7} \\
            distill & 1024 & 200K & 18.11 \\ 
            \hline
            from scratch & 1024 & 337k & \textbf{26.94} \\
            distill & 1024 & 337K & 25.44 \\ 
            \hline
            from scratch & 256 & 1M & 27.36 \\
            distill & 256 & 1M & \textbf{28.12} \\ \bottomrule
	\end{tabular}
\end{table}

\balance


\section{Conclusion}

Our work demonstrates the potential of combining knowledge distillation and quantization to develop efficient, deployable multi-task RL agents, significantly reducing model size while maintaining or improving performance. However, key limitations remain, including the need for real-world deployment on hardware platforms, the reliance on high-capacity teacher models, and the narrow focus on the MT30 benchmark. Despite these challenges, our approach lays a foundation for scaling RL models to resource-constrained environments, paving the way for more accessible and practical multi-task RL systems.





\bibliographystyle{ACM-Reference-Format} 
\bibliography{main}


\begin{thebibliography}{8}


\ifx \showCODEN    \undefined \def \showCODEN     #1{\unskip}     \fi
\ifx \showDOI      \undefined \def \showDOI       #1{#1}\fi
\ifx \showISBNx    \undefined \def \showISBNx     #1{\unskip}     \fi
\ifx \showISBNxiii \undefined \def \showISBNxiii  #1{\unskip}     \fi
\ifx \showISSN     \undefined \def \showISSN      #1{\unskip}     \fi
\ifx \showLCCN     \undefined \def \showLCCN      #1{\unskip}     \fi
\ifx \shownote     \undefined \def \shownote      #1{#1}          \fi
\ifx \showarticletitle \undefined \def \showarticletitle #1{#1}   \fi
\ifx \showURL      \undefined \def \showURL       {\relax}        \fi
\providecommand\bibfield[2]{#2}
\providecommand\bibinfo[2]{#2}
\providecommand\natexlab[1]{#1}
\providecommand\showeprint[2][]{arXiv:#2}

\bibitem[\protect\citeauthoryear{Czarnecki, Pascanu, Osindero, Jayakumar,
  Swirszcz, and Jaderberg}{Czarnecki et~al\mbox{.}}{2019}]%
        {czarnecki2019distilling}
\bibfield{author}{\bibinfo{person}{Wojciech~M Czarnecki},
  \bibinfo{person}{Razvan Pascanu}, \bibinfo{person}{Simon Osindero},
  \bibinfo{person}{Siddhant Jayakumar}, \bibinfo{person}{Grzegorz Swirszcz},
  {and} \bibinfo{person}{Max Jaderberg}.} \bibinfo{year}{2019}\natexlab{}.
\newblock \showarticletitle{Distilling policy distillation}. In
  \bibinfo{booktitle}{\emph{The 22nd International Conference on Artificial
  Intelligence and Statistics}}. PMLR, \bibinfo{pages}{1331--1340}.
\newblock


\bibitem[\protect\citeauthoryear{Hansen, Su, and Wang}{Hansen
  et~al\mbox{.}}{2024}]%
        {hansen2024td}
\bibfield{author}{\bibinfo{person}{Nicklas Hansen}, \bibinfo{person}{Hao Su},
  {and} \bibinfo{person}{Xiaolong Wang}.} \bibinfo{year}{2024}\natexlab{}.
\newblock \showarticletitle{TD-MPC2: Scalable, Robust World Models for
  Continuous Control}.
\newblock \bibinfo{journal}{\emph{arXiv preprint arXiv:2310.16828}}
  (\bibinfo{year}{2024}).
\newblock


\bibitem[\protect\citeauthoryear{Hinton, Vinyals, and Dean}{Hinton
  et~al\mbox{.}}{2015}]%
        {hinton2015distilling}
\bibfield{author}{\bibinfo{person}{Geoffrey Hinton}, \bibinfo{person}{Oriol
  Vinyals}, {and} \bibinfo{person}{Jeff Dean}.}
  \bibinfo{year}{2015}\natexlab{}.
\newblock \showarticletitle{Distilling the knowledge in a neural network}. In
  \bibinfo{booktitle}{\emph{NIPS Deep Learning and Representation Learning
  Workshop}}.
\newblock


\bibitem[\protect\citeauthoryear{Micikevicius, Narang, Alben, Diamos, Elsen,
  Garcia, Ginsburg, Houston, Kuchaiev, Venkatesh, et~al\mbox{.}}{Micikevicius
  et~al\mbox{.}}{2018}]%
        {micikevicius2018mixed}
\bibfield{author}{\bibinfo{person}{Paulius Micikevicius},
  \bibinfo{person}{Sharan Narang}, \bibinfo{person}{Jonah Alben},
  \bibinfo{person}{Gregory Diamos}, \bibinfo{person}{Erich Elsen},
  \bibinfo{person}{David Garcia}, \bibinfo{person}{Boris Ginsburg},
  \bibinfo{person}{Michael Houston}, \bibinfo{person}{Oleksii Kuchaiev},
  \bibinfo{person}{Ganesh Venkatesh}, {et~al\mbox{.}}}
  \bibinfo{year}{2018}\natexlab{}.
\newblock \showarticletitle{Mixed precision training}.
\newblock \bibinfo{journal}{\emph{arXiv preprint arXiv:1710.03740}}
  (\bibinfo{year}{2018}).
\newblock


\bibitem[\protect\citeauthoryear{Parisotto, Ba, and Salakhutdinov}{Parisotto
  et~al\mbox{.}}{2015}]%
        {parisotto2015actor}
\bibfield{author}{\bibinfo{person}{Emilio Parisotto},
  \bibinfo{person}{Jimmy~Lei Ba}, {and} \bibinfo{person}{Ruslan
  Salakhutdinov}.} \bibinfo{year}{2015}\natexlab{}.
\newblock \showarticletitle{Actor-mimic: Deep multitask and transfer
  reinforcement learning}.
\newblock \bibinfo{journal}{\emph{arXiv preprint arXiv:1511.06342}}
  (\bibinfo{year}{2015}).
\newblock


\bibitem[\protect\citeauthoryear{Rusu, Colmenarejo, Gulcehre, Desjardins,
  Kirkpatrick, Pascanu, Mnih, Kavukcuoglu, and Hadsell}{Rusu
  et~al\mbox{.}}{2015}]%
        {rusu2015policy}
\bibfield{author}{\bibinfo{person}{Andrei~A Rusu},
  \bibinfo{person}{Sergio~Gomez Colmenarejo}, \bibinfo{person}{Caglar
  Gulcehre}, \bibinfo{person}{Guillaume Desjardins}, \bibinfo{person}{James
  Kirkpatrick}, \bibinfo{person}{Razvan Pascanu}, \bibinfo{person}{Volodymyr
  Mnih}, \bibinfo{person}{Koray Kavukcuoglu}, {and} \bibinfo{person}{Raia
  Hadsell}.} \bibinfo{year}{2015}\natexlab{}.
\newblock \showarticletitle{Policy distillation}.
\newblock \bibinfo{journal}{\emph{arXiv preprint arXiv:1511.06295}}
  (\bibinfo{year}{2015}).
\newblock


\bibitem[\protect\citeauthoryear{Tassa, Doron, Muldal, Erez, Li, Casas, Budden,
  Abdolmaleki, Merel, Lefrancq, et~al\mbox{.}}{Tassa et~al\mbox{.}}{2020}]%
        {tassa2020dmcontrol}
\bibfield{author}{\bibinfo{person}{Yuval Tassa}, \bibinfo{person}{Yotam Doron},
  \bibinfo{person}{Alistair Muldal}, \bibinfo{person}{Tom Erez},
  \bibinfo{person}{Yazhe Li}, \bibinfo{person}{Diego de~Las Casas},
  \bibinfo{person}{David Budden}, \bibinfo{person}{Abbas Abdolmaleki},
  \bibinfo{person}{Josh Merel}, \bibinfo{person}{Andrew Lefrancq},
  {et~al\mbox{.}}} \bibinfo{year}{2020}\natexlab{}.
\newblock \showarticletitle{Dm\_control: Software and tasks for continuous
  control}.
\newblock \bibinfo{journal}{\emph{arXiv preprint arXiv:2006.12983}}
  (\bibinfo{year}{2020}).
\newblock


\bibitem[\protect\citeauthoryear{Teh, Bapst, Czarnecki, Quan, Kirkpatrick,
  Hadsell, Heess, and Pascanu}{Teh et~al\mbox{.}}{2017}]%
        {teh2017distral}
\bibfield{author}{\bibinfo{person}{Yee~Whye Teh}, \bibinfo{person}{Victor
  Bapst}, \bibinfo{person}{Wojciech~M Czarnecki}, \bibinfo{person}{John Quan},
  \bibinfo{person}{James Kirkpatrick}, \bibinfo{person}{Raia Hadsell},
  \bibinfo{person}{Nicolas Heess}, {and} \bibinfo{person}{Razvan Pascanu}.}
  \bibinfo{year}{2017}\natexlab{}.
\newblock \showarticletitle{Distral: Robust multitask reinforcement learning}.
  In \bibinfo{booktitle}{\emph{Advances in Neural Information Processing
  Systems}}. \bibinfo{pages}{4496--4506}.
\newblock


\end{thebibliography}


\end{document}